\documentclass{article}

\usepackage{microtype}
\usepackage{graphicx}
\usepackage{booktabs} 
\usepackage{hyperref}
\usepackage{amsmath}
\usepackage{amssymb}
\usepackage{graphics}
\usepackage{graphicx}
\usepackage{subfig}
\usepackage{listings}
\usepackage{fullpage}
\usepackage{authblk}

\def\R{{\mathbb R}}

\def\sumi{\sum_{i=1}^n} 
\def\sumj{\sum_{j=1}^n}

\def\bp{\noindent {\it Proof.}\ }
\def\ep{\hfill $\Box$} 

\def\be{\begin{equation}}
\def\ee{\end{equation}}

\def\diag{\text{\rm diag}}

\newtheorem{prop}{Proposition}

\begin{document}

\title{Weighted Spectral  Embedding of Graphs}
\author[1]{Thomas Bonald}
\author[2]{Alexandre Hollocou}
\author[2]{Marc Lelarge}
\affil[1]{Telecom ParisTech, Paris, France}
\affil[2]{Inria, Paris, France}
\date{\today}

\maketitle

\begin{abstract}
We present a novel spectral embedding of  graphs that incorporates   weights assigned to the nodes, quantifying their  relative importance.
This  spectral embedding  is based on the first eigenvectors of some  properly normalized  version of the Laplacian.
 We prove that these eigenvectors correspond to the  configurations of lowest energy of an equivalent physical system, either mechanical or electrical, in which the weight of each node  can be interpreted as its mass or its capacitance, respectively.  
Experiments on a real dataset illustrate the impact of  weighting on the  embedding.
\end{abstract}

\section{Introduction}

Many types of data can be represented as graphs. Edges may correspond  to actual links in the data (e.g., users connected by some social network) or to  levels of similarity induced from the data (e.g., users having liked a large common set  of  movies). The resulting graph is typically {\it sparse} in the sense  that the number of edges  is much lower than  the total number of node pairs, which makes the data hard to exploit. 

A standard approach to the analysis of sparse graphs  consists in {\it embedding}  the graph in some vectorial space of  low dimension, typically much smaller than the number of nodes \cite{robles2007riemannian,yan2007graph,cai2018comprehensive}. Each node is represented by some vector in the embedding space so that  
 close nodes in the graph (linked either directly or through many short paths in the graph) tend to be represented  by close vectors in terms of  the Euclidian distance. Standard learning techniques can then be applied to this dense  vectorial representation of the graph to recommend new links, rank nodes or find clusters of nodes for instance \cite{fouss2007random,bronstein2017geometric}.

The most popular technique for graph  embedding is based on the spectral decomposition of the graph Laplacian, each dimension of the embedding space corresponding to an eigenvector of the Laplacian matrix \cite{chung,ng2002spectral,belkin2003laplacian,luxburg,spielman2007spectral,newman13}. 
This  spectral embedding can be interpreted in terms of a random walk in the graph \cite{lovasz93,qiu2007clustering}. In the full embedding space (including all eigenvectors),  the square distance between two vectors  is proportional to the {\it mean commute time} of the random walk between the two corresponding nodes: close nodes in the graph tend to be  close in the embedding space. Viewing this random walk as the path followed by electrons in the corresponding electrical network, with nodes  linked by resistors,
these square distances can also be interpreted as the  {\it effective resistances} between  pairs of  nodes   \cite{snell00}.

In this paper, we address the issue of the spectral embedding of graphs including {\it node weights}. We shall see that existing spectral embedding techniques implicitly consider either unit weights or so-called internal node weights, depending on the Laplacian used in the spectral decomposition. We here consider the node weights as some additional information representing the relative importance of the nodes, independently of the graph structure. The weight of a node  may   reflect either its {\it value},   its {\it multiplicity}   if each node represents a category of   users or items,  or the {\it reliability} of the associate data,  for instance. Surprisingly,  this notion of weight is common for vectoral data (see, e.g.,  the weighted version of the  k-means clustering algorithm  \cite{huang1998extensions,dhillon2004kernel,kerdprasop2005weighted}) but not for graph data, where weights are typically  assigned to edges but  not to nodes, apart from those induced from the edges.

Our main contribution  is a spectral embedding of the graph, we refer to as the {\it weighted} spectral embedding, that incorporates the node weights. It is based on the spectral decomposition of some properly normalized version of the Laplacian. We prove that, when all eigenvectors are used, this embedding is equivalent to the regular spectral embedding  {\it shifted} so that the origin is the center of mass of the embedding. In practice, only the first eigenvectors are included to get an embedding in low dimension. 
We show that these eigenvectors can be interpreted as the  levels of lowest energy of a physical system, either a mechanical system where nodes are linked by {\it springs} (the edges) and have different {\it masses} (the node weights), or an electrical network where nodes are linked by {\it resistors} (the edges)  and connected to the ground by capacitors with different {\it capacitances}  (the node weights). 
In particular, the  weighted spectral embedding  can {\it not} be  derived from the regular spectral embedding in low dimension. Experiments 
confirm that these embeddings differ significantly in practice.

The weighted spectral embedding can also be interpreted in terms of a random walk in the graph, where nodes are visited in proportion to their weights.
In the full embedding space (including all eigenvectors),  the square distances between  vectors  are proportional to the {\it mean commute times} of this random walk between the  corresponding nodes, as for unit weights.
In fact, these 
   mean commute times depend on the weights through their sum only, which explains  why  the pairwise distances of the embedding are independent of the weights, up to some multiplicative constant.      This property is somewhat counter-intuitive as the mean hitting time of one node from another {\it does} depend on the weights. We shall explain this apparent paradox  by some symmetry property of each equivalent physical system.

The rest of the paper is organized as follows. We first introduce the model and the notations. We then present  the regular  spectral embedding and its interpretation in terms of a random walk in the graph. The weighted version of this random walk  is introduced in Section \ref{sec:walk}. Section \ref{sec:weighted} presents the weighted spectral embedding and extends the results known for the regular spectral embedding. The analogies with a mechanical system and an electrical network are described in Sections \ref{sec:mech} and \ref{sec:elec}, respectively.  Experiments on real data are presented in Section \ref{sec:exp}. Section \ref{sec:conc} concludes the paper.



\section{Model}
\label{sec:graph}

We consider a connected, undirected graph of $n$ nodes, without self-loops. We denote by $A$ its adjacency  matrix. In the absence of edge weights, this is a binary, symmetric matrix, with 
$A_{ij} = 1$ if and only if there is an edge between nodes $i$ and $j$.  In the presence of edge weights, $A_{ij}$ is the weight of the edge between nodes $i$ and $j$, if any, and is equal to 0 otherwise. 

Let $d= Ae$, where $e$ is the $n$-dimensional  vector of ones. The components $d_1,\ldots,d_n$ of the vector $d$ are equal to the actual node degrees in the absence of edge weights ($A$ is a binary matrix) and to the total weights  of incident edges otherwise ($A$ is a  non-negative matrix).   
We refer to $d_1,\ldots,d_n$  as the  {\it internal} node weights. 

Nodes are assigned positive weights $w_1,\ldots,w_n$ corresponding to their relative importances. These node weights are external parameters, independent of the graph. We denote by $w$ the vector $(w_1,\ldots,w_n)$.

\section{Spectral embedding}
\label{sec:spectral}

We first present the regular spectral embedding, without taking the node weights $w_1,\ldots,w_n$ into account.
Let $D = \diag(d)$ be the diagonal matrix of internal node weights. The  Laplacian matrix is defined by
$$
L = D-A.
$$
This  is a  symmetric matrix. It is  positive semi-definite on observing that:
$$
\forall v\in \R^n,\quad v^TLv = \sum_{i<j} A_{ij}(v_j - v_i)^2.
$$
 The spectral theorem yields
\be\label{eq:spectral}
 L =  U\Lambda U^T,
\ee
where $\Lambda= \diag(\lambda_1,\ldots,\lambda_n)$ is the diagonal matrix of eigenvalues of $L$, with $0 =  \lambda_1 < \lambda_2  \le \ldots \le \lambda_n$, and $U=(u_1,\ldots,u_n)$ is  the matrix of  corresponding eigenvectors, with $U^TU=I$ and $u_1 = e / \sqrt{n}$.

\paragraph*{Spectral embedding.} Let $X = \sqrt{\Lambda^+} U^T$, where $\Lambda^+ =  \diag(0, 1 /{\lambda_2},\ldots,1/\lambda_n)$ denotes the pseudo-inverse of $\Lambda$. The columns $x_1,\ldots,x_n$ of the matrix $X$ define an embedding of the nodes in $\R^n$, each dimension corresponding to an eigenvector of the Laplacian.  Observe that the first component of each vector $x_1,\ldots,x_n$ is equal to 0, reflecting the fact that the first eigenvector is not informative. Since $Xe = 0$,  the centroid of the $n$ vectors is the origin:
\be\label{eq:barycentre}
\frac 1 n \sumi x_i = 0.
\ee
The Gram matrix of $X$ is the pseudo-inverse of the Laplacian:
$$X^TX = U\Lambda^+ U^T  = L^+.$$
 
\paragraph*{Random walk.}
Consider a random walk in the graph where the transition rate from node $i$ to node $j$ is $A_{ij}$. Specifically, the walker stays at node $i$ an exponential time with parameter $d_i$, then moves  from node $i$ to node $j$ with probability $P_{ij} = A_{ij}/d_i$. This defines a continuous-time Markov chain   with generator matrix $-L$ and  uniform stationary  distribution. 
The sequence of  nodes   visited by the  random walk forms a discrete-time Markov  chain  with transition 
matrix $P= D^{-1} A$.

\paragraph*{Hitting times.}
Let  $H_{ij}$   be the mean hitting time  of node $j$ from node $i$. Observe that $H_{ii}=0$.
The  following results, proved in \cite{lovasz93,qiu2007clustering}, will be extended to  weighted spectral embedding in Section \ref{sec:walk}.
We denote by $e_i$ the $n$-dimensional unit vector on component $i$.

\begin{prop}
The matrix $H$ satisfies:
\be\label{eq:hit-pb1}
L  H = e e^T -nI.
\ee
\end{prop}

\begin{prop}\label{prop:hit1}
The  solution $H$  to equation \eqref{eq:hit-pb1} with zero diagonal entries  satisfies:
\be\label{eq:hit-sol1}
\forall i,j,\quad H_{ij} = n(e_j - e_i)^T L^+ e_j. 
\ee
\end{prop}


\vspace{.5cm}

Since $L^+ = X^TX$, we obtain:
$$
H_{ij} = n(x_j-x_i)^Tx_j.
$$
We deduce the  mean commute time  between nodes $i$ and $j$:
$$
C_{ij} = H_{ij} + H_{ji} =  n ||x_i -x_j||^2.
$$
In view of \eqref{eq:barycentre}, the mean hitting time of node $j$ from steady state is:
$$
h_j = \frac 1 n \sumi H_{ij} =n  ||x_j||^2.
$$

\paragraph*{Cosine similarity.}
Observe that:
$$
n x_i^T x_j = h_j - H_{ij}  = h_i - H_{ji}.
$$
Let
$
S_{ij} = \cos(x_i,x_j).
$
This is the cosine-similarity between nodes $i$ and $j$.
We have:
$$
S_{ij} = \frac{h_j - H_{ij}}{\sqrt{h_ih_j}} =  \frac{h_i - H_{ji}}{\sqrt{h_ih_j}},
$$
that is
\be\label{eq:relative}
S_{ij} = \frac{h_i + h_j - C_{ij}}{2\sqrt{h_ih_j}}.
\ee
Thus the cosine-similarity between any vectors $x_i,x_j$  can be interpreted in terms of the mean commute time between the corresponding nodes $i,j$ {\it relative} to their mean hitting times.


\section{Random walk with weights}
\label{sec:walk}

In this section, we introduce a modified version of the random walk, that takes the  node weights into account. Recall that all weights are assumed positive. We denote by  $|w| = \sum_{i=1}^n w_i$ the total node weight.

\paragraph*{Random walk.}
We modify the random walk as follows: the transition rate from node $i$ to node $j$ is now $A_{ij}/ w_i$. Thus the walker stays at node $i$ an exponential time with parameter $d_i/w_i$, then moves  from node $i$ to node $j$ with probability $P_{ij}$. Observe that the previously considered  random walk corresponds to unit weights.
We get a continuous-time Markov chain   with generator matrix $-W^{-1}L$, with $W = \diag(w)$, and   stationary  distribution $\pi = w / |w|$.

\paragraph*{Hitting times.}
Let  $H_{ij}$   be the mean hitting time  of node $j$ from node $i$. Observe that $H_{ii}=0$.
\begin{prop}
The matrix $H$ satisfies:
\be\label{eq:hit-pb}
L  H = {w} e^T - |w| I.
\ee
\end{prop}
\bp
We have $H_{ii} = 0$ while for all $i\ne j$,
$$ 
H_{ij} = 
\frac {w_i}{d_i}+ \sum_{k=1}^n P_{ik} H_{kj}.
$$
Thus the matrix $(I - P)H - D^{-1}w e^T$ is diagonal. Equivalently, 
the matrix 
$ L H  - w e^T
$ 
is diagonal. Since $e^T  L= 0$,  this diagonal matrix  is 
$- |w|I$. \ep

\begin{prop}\label{prop:hit}
The  solution $H$  to equation \eqref{eq:hit-pb} with zero diagonal entries  satisfies:
$$
\forall i,j,\quad H_{ij} = |w|(e_j - e_i)^T L^+ (e_j-\pi). 
$$
\end{prop}
\bp
The matrix $M = w  e^T - |w| I$ satisfies $e^T M  = 0$ so the seeked solution  is of the form:
$$
H =   L^+ (w e^T - |w| I) + e h^T,
$$
for some $n$-dimensional vector $h$. Thus,
$$
H_{ij} = |w| e_i^T L^+  (\pi - e_j) + h_j.
$$
Since $H_{jj} = 0$, we get:
$$
h_j = -|w| e_j^T L^+ (\pi - e_j).
$$
\ep
\\

Let $\bar x$ be the center of mass of the vectors $x_1,\ldots,x_n$:
$$
\bar x = X \pi =  \sumi \pi_i x_i.
$$
Since $L^+ = X^TX$, we obtain:
\be\label{eq:Hij}
H_{ij} = |w|(x_j-x_i)^T(x_j - \bar x).
\ee
Thus the mean hitting times are the {\it same} as without node weights, up to the multiplicative constant $|w|$ and the shift of the origin to the center of mass $\bar x$ of the vectors $x_1,\ldots,x_n$.

In particular, the  mean  commute time   between any nodes $i$ and $j$ depends on the weights through their sum only:
$$
C_{ij} = H_{ij} + H_{ji} =  |w| ||x_i -x_j||^2.
$$
The mean hitting time of node $j$ from steady state is:
$$
h_j =  \sumi \pi_i H_{ij} =|w|  ||x_j- \bar x||^2.
$$


\paragraph*{Cosine similarity.}
Let $
S_{ij} = \cos(x_i - \bar x,x_j - \bar x).
$
This is the cosine-similarity between nodes $i$ and $j$. 
We obtain as above:
$$
S_{ij} = \frac{h_i + h_j - C_{ij}}{2\sqrt{h_ih_j}}.
$$
In particular, the relative mean commute times, as defined by \eqref{eq:relative},  depend on the weights $w_1,\ldots,w_n$ through the center of mass $\bar x$ of the vectors $x_1,\ldots,x_n$ only.

\section{Weighted spectral embedding}
\label{sec:weighted}

We now introduce the weighted spectral embedding. 
The  generator matrix $-W^{-1}L$ of the weighted random walk  is not symmetric in general. Thus we consider the following normalized version of the  Laplacian, we refer to as the {\it weighted} Laplacian:
$$
 L_W  = W^{-\frac 1 2} L W^{-\frac 1 2}.
$$
This matrix is   symmetric, positive semi-definite. 
In the particular case where $W = D$, this is known as the symmetric  normalized   Laplacian.

The spectral theorem yields
\be\label{eq:spectral}
L_W = \hat U \hat \Lambda \hat U^T,
\ee
where $\hat \Lambda= \diag(\hat\lambda_1,\ldots,\hat\lambda_n)$ is the diagonal matrix of eigenvalues of $L_W$, with $0 = \hat \lambda_1 < \hat \lambda_2  \le \ldots \le \hat \lambda_n$, and $\hat U=(\hat u_1,\ldots,\hat u_n)$ is  the matrix of  corresponding eigenvectors, with $\hat U^T\hat U=I$ and $\hat u_1 = \sqrt{\pi}$. 

We have the  following explicit expression for the pseudo-inverse of the weighted Laplacian:

\begin{prop}\label{prop:pseudow}
The pseudo-inverse of $L_W$ is:
\be\label{eq:pseudoW}
 L_W^+  = W^{\frac 1 2}(I - e \pi^T) L^+ (I - \pi e^T)W^{\frac 1 2}.
\ee
\end{prop}
\bp
Let $M$ be the matrix defined by the right-hand side of  \eqref{eq:pseudoW}.
Using the fact that $Le = 0$, we get:
\begin{align*}
L_W M  L_W& = W^{-\frac 1 2} L L^+ L W^{-\frac 1 2},\\
&  =  W^{-\frac 1 2} L^+  W^{-\frac 1 2}=L_W,
\end{align*}
which proves that $M$ is  the pseudo-inverse of $L_W$.
\ep
\\

\paragraph*{Generalized eigenvalue problem.}
Let $V = W^{-\frac 1 2} \hat U$. We have:$$LV = WV \hat \Lambda,$$
with $V^T W V = I$.
Thus the columns $v$ of the matrix $V$ are  solutions to the generalized eigenvalue problem:
\be\label{eq:geig}
L v = \lambda W v,
\ee
with corresponding eigenvalues $\lambda = \hat \lambda_1,    \ldots, \hat \lambda_n.$
The first column, associated to the eigenvalue $\hat \lambda_1 = 0$,  satisfies $v\propto e$, while the others satisfy $w^T v = 0$.

\paragraph*{Spectral embedding.} Let $Y = \sqrt{\Lambda^+} \hat U^T W^{-\frac 1 2}$. The columns $y_1,\ldots,y_n$ of the matrix $Y$ define an embedding of the nodes, we refer to as the {\it weighted} spectral embedding.  As for the regular spectral embedding, the first component of each vector  $y_1,\ldots,y_n$  is equal to 0.
Since $Yw = 0$, the center of mass of $y_1,\ldots,y_n$ lies at the origin:
$$
\sum_{i=1}^n w_i y_i = 0.
$$

The Gram matrix of $Y$ is:
$$Y^TY = W^{-\frac 1 2} \hat U\hat \Lambda^+ \hat U^T  W^{-\frac 1 2} = W^{-\frac 1 2} L_W^+ W^{-\frac 1 2}.$$

In view of Proposition \ref{prop:pseudow},
\begin{align*}
y_i^T y_j &=  e_i^T W^{-\frac 1 2} L_W^+ W^{-\frac 1 2} e_j,\\
&= (e_i - \pi)^T L^+ (e_j - \pi)^T,\\
& = (x_i-\bar x)^T(x_j - \bar x).
\end{align*}
Thus the considered weighted spectral embedding is equivalent to the regular spectral embedding, shifted so that the origin is  the center of mass   $\bar x$.
In particular, the distances between vectors in the embedding space can be interpreted in terms of mean hitting times of the random walk, as shown in Section \ref{sec:walk}.

\section{A mechanical system}
\label{sec:mech}

Consider the mechanical system consisting of  $n$ point particles of respective masses $w_1,\ldots,w_n$ sliding along a bar without friction.
Particles $i$ and $j$ are linked by a spring  satisfying Hooke's law with stiffness $A_{ij}$.

\paragraph*{Eigenmodes.}
Assume that the bar has a uniform circular motion  with angular velocity $\omega$ around some fixed axis. We denote by  $v_1,\ldots,v_n$ the locations of the particles along the bar, with the axis taken as the origin.  By Newton's second law of motion, the system is in equilibrium if and only if 
$$
\forall i=1,\ldots,n, \quad \sumj A_{ij} (v_j - v_i) = -  w_i v_i \omega^2,
$$
that is 
$$
Lv = \omega^2 W v.
$$
If $v \ne 0$, then $v$ is a solution to  the generalized eigenvector problem \eqref{eq:geig} with eigenvalue $\lambda = \omega^2$. 
We refer to these equilibrium states as the {\it eigenmodes} of the mechanical system.

The first eigenmode $v\propto e$, for $\omega^2 = 0$, corresponds to the absence of motion. The other eigenmodes give the possible angular velocities of the system, equal to the square roots of the eigenvalues $\hat \lambda_2,\ldots,\hat \lambda_n$ of $L_W$.
 Any such  eigenmode $v$ satisfies $w^T v = 0$, meaning that 
 the center of mass of the system is  at the origin.

\paragraph*{Potential energy.}
The mechanical potential energy of the system in any state $v$ is:
$$
\frac 1 2 v^TLv.  
$$
If  $v$ is an eigenmode of the mechanical system, this is equal to:
$$
 \frac \lambda 2  v^TWv.
$$
As 
  $v^TWv$ is the moment of inertia of the system and $\lambda = \omega^2$ is the square of the angular velocity, this  corresponds to the angular kinetic energy. For a unit moment of inertia  $v^TWv=1$, we obtain
$ 
 \lambda = v^TLv,
 $
so that the eigenvalues of the weighted Laplacian $L_W$ can be viewed as (twice) the levels of  energy of the eigenmodes, for unit moments of inertia.
The weighted spectral embedding reduced to the $k$ first eigenvectors of $L_W$  can then 
be interpreted as that induced by   the $k$ eigenmodes of lowest potential energies  of the mechanical system, for unit moments of inertia.


\paragraph*{Dirichlet problem.}
For any  $i\ne j$, assume the positions of the point particles $i$ and $j$ are set to 1 and 0, respectively. 
 Let $v$ be the vector of positions at equilibrium, in the absence of rotation. By Newton's first law of motion,
 $$
 Lv=  \alpha(e_i - e_j),
 $$
 for some constant $\alpha$ equal to the force $F$ exerted on both   $i$ and $j$ (in opposite directions). This force does {\it not} depend on the masses of the point particles.

The solution to this Dirichlet problem is:
 $$
 v = \alpha L^+(e_i - e_j) + \beta e,
 $$
 for some constant $\beta$. Using the fact that $v_i = 1$ and  $v_j = 0$, we get:
 $$
 v_k = \frac{(e_k- e_j)^T L^+ (e_i - e_j)}{(e_i- e_j)^T L^+ (e_i - e_j)} = \frac{(x_k - x_j)^T(x_i - x_j)}{||x_i - x_j||^2}
 $$
 and
 $$
 \alpha = \frac 1 {||x_i - x_j||^2}.
 $$
 Observe in particular that $v_k \in [0,1]$ for all nodes $k$.
The mean hitting time of node $j$ from node $i$ by the random walk, given by \eqref{eq:Hij}, satisfies:
$$
H_{ij} = \frac {|w|} F \bar v,
$$
where $\bar v$ is the center of mass of the system:
$$
\bar v =\sum_{k=1}^n \pi_k v_k.
$$
By symmetry, we have:
 $$
 H_{ji} =  \frac{|w|}{F}(1-\bar v).
 $$
 We deduce the mean commute time between nodes $i$ and $j$:
 $$
 C_{ij} = H_{ij} +  H_{ji} =\frac{|w|}{F}.
 $$
  This symmetry in the solutions to the Dirichlet problem explains why, unlike the mean hitting times, the mean commute times depend on the weights through their sum only.
 The mean commute time between two nodes of the graph is inversely proportional to  the force exerted between the two corresponding point particles  of the mechanical system, which is independent of the particle masses.

 %
%
%
%
%

\section{An electrical network}
\label{sec:elec}

Consider the electrical network induced by the graph, with a resistor of conductance $A_{ij}$ between nodes $i$ and $j$ and a capacitor of capacitance $w_i$ between node $i$ and the ground. 
The move of electrons in the network is precisely described by  the random walk defined in Section \ref{sec:walk}.

\paragraph*{Eigenmodes.}
Let $v(t)$ be the vector of electric potentials at time $t$, starting from some arbitrary initial state $v$. By Ohm's law,
$$
\forall i=1,\ldots,n, \quad  w_i \frac{dv_i(t)}{dt}  = \sumj A_{ij} (v_j(t) - v_i(t)),
$$
that is
$$
W\frac{dv(t)}{dt} = - L v(t).
$$
The solution to this differential equation is:
$$
v(t) =  e^{-W^{-1}Lt} v.
$$
If the initial state $v$ is a solution to the generalized eigenvector problem \eqref{eq:geig}, we obtain:
$$
v(t) =   e^{-\lambda t} v.
$$
Since  the eigenvectors of $L_W$ form a basis of $\R^n$, 
any initial state $v$ can be written as a linear combination of  solutions to the generalized eigenvector problem, we refer to as the {\it eigenmodes} of the electrical network.
The first eigenmode $v  \propto e$, for $\lambda = 0$, corresponds to a static system, without  discharge.
The other eigenmodes give the possible discharge rates of the system, equal to  the  eigenvalues $\hat \lambda_2,\ldots,\hat \lambda_n$ of $L_W$.
 Any such  eigenmode $v$ satisfies $w^T v = 0$, meaning that the total charge accumulated in the capacitors is null.

\paragraph*{Energy dissipation.}
The energy dissipated by the resistors for the vector of electric potentials $v$ is:
$$
 \frac 1{2} v^TLv.
$$ 
If  $v$ be an eigenmode of the electrical network, this is equal to
$$
\frac \lambda 2  v^T W  v.
$$
Observing that $\frac 1 2 v^T W  v$ is the 
 electrical potential energy of the system, we deduce that  $\lambda$ can be viewed as   the energy dissipated by the resistors   for a unit electric potential energy. 
In particular, the weighted spectral embedding reduced to  the $k$ first eigenvectors of $L_W$ can be interpreted as that induced by   the $k$ eigenmodes of lowest energy dissipation of the electrical network, for unit electric potential energies.  

%

\paragraph*{Dirichlet problem.}
For any  $i\ne j$, assume the electric potentials of nodes $i$ and $j$ are set to 1 and 0, respectively. 
 Let $v$ be the vector of electric potentials at equilibrium. We have:
 $$
 Lv=  \alpha(e_i - e_j),
 $$
 for some constant $\alpha$ equal to the current flowing from  $i$ to $j$. 
  This  Dirichlet problem  is the same as for the mechanical network, and thus the same results apply. 
 In particular, we get $v_k\in [0,1]$ for all nodes $k$, and the current  between   $i$ to $j$ is:
 $$
 \alpha = \frac 1 { ||x _i - x_j||^2}.
 $$
 This is the intensity of the current  $I$ generated by a unit potential difference   betweeen $i$ and $j$.
Its inverse  is known as the {\it effective resistance} between $i$ and $j$.
 The mean hitting time of node $j$ from node $i$ by the random walk satisfies:
$$
H_{ij} = \frac  q I,
$$
where $q$ is
 the total charge accumulated in the capacitors:
$$
q = \sum_{k=1}^n w_k v_k. 
$$
Observe that this expression for the mean hitting time can be seen as a consequence of Little's law, as $I$ is the intensity of positive charges entering the network and $q$ the mean number of positive charges accumulated in the network.

By symmetry, the total charge accumulated in the capacitors when the respective electric potentials of $i$ and $j$ are set to 0 and 1 is:
 $$
   \sum_{k=1}^n w_k (1 - v_k) = |w|  - q,
 $$
so that:
 $$
 H_{ji} =  \frac{|w| - q} I.
 $$
 We deduce the mean commute time between nodes $i$ and $j$:
 $$
 C_{ij} = H_{ij} +  H_{ji} = \frac{|w|} I.
 $$
 Again, the  symmetry in the solutions to the Dirichlet problem explains why the mean commute times depend on the weights through their sum only.
The mean commute time between two nodes of the graph is  proportional to  the effective resistance between the two corresponding points   of the  electrical network, which is independent  of the capacitors.

\section{Experiments}
\label{sec:exp}

We now illustrate the results on  a real dataset.  
The considered graph  is that formed by  articles of Wikipedia for Schools\footnote{\url{https://schools-wikipedia.org}}, a selection of  articles of Wikipedia for children   \cite{west2009}.
Specifically, we have extracted the largest connected component of this graph, considered as undirected. The resulting graph has 4,589 nodes (the articles) and 106,644 edges (the hyperlinks between  these articles). The graph is undirected and unweighted. Both the dataset and the Python code used for the experiments are available online\footnote{\url{https://github.com/tbonald/spectral_embedding}}.

\paragraph*{Global clustering.}
We first apply   k-means clustering\footnote{The algorithm is k-means++ \cite{arthur2007k}  100 random initializations.} to the following embeddings of the graph, each  in dimension $k = 100$:
\begin{itemize}
\item the regular spectral embedding, $X$, restricted to rows $2,\ldots,k + 1$, say $\hat X$;
\item the shifted spectral embedding, $\hat X - \hat X\pi 1^T$, where $\pi = w / |w|$ and $1$ is the $k$-dimensional vector of ones;
\item the weighted spectral embedding, $Y$, restricted to rows $2,\ldots,k + 1$, say $\hat Y$.
\end{itemize}
Here the vector of weights $w$ is taken equal to $d$, the vector of internal node weights. In particular, the weighted spectral embedding is that following from the spectral decomposition of the  normalized Laplacian, $D^{-\frac 1 2} L D^{-\frac 1 2}$.
Observe that the first row of each embedding $X$ and $Y$  is equal to zero and thus discarded.

Each  embedding is normalized so that  nodes are represented by  $k$-dimensional unitary vectors. This is equivalent to consider the distance induced by the {\it cosine similarity} in the original embedding. In particular, the regular spectral embedding and the shifted spectral embedding give different clusterings.

Tables \ref{tab:global3} and  \ref{tab:global1} show the top  articles of the  clusters found for each embedding, when the number of clusters is set to 20, with the size of each cluster. The selected articles for each cluster correspond to the nodes of highest degrees among the $50\%$ closest  nodes from the center of mass of the cluster in the embedding space, with unit weights for the regular spectral embedding and internal node weights $d$ for the  shifted spectral embedding and the weighted spectral embedding.

 \begin{table}[h]
\begin{center}
\begin{tabular}{|c|l|}
\hline
 Size & Top articles\\
\hline
1113 & Australia, Canada, North America, 20th century\\
326 & UK, England, London, Scotland, Ireland\\
250 & US, New York City, BBC, 21st century, Los Angeles\\
227 & India, Japan, China, United Nations\\
218 & Earth, Sun, Physics, Hydrogen, Moon, Astronomy\\
200 & Mammal, Fish, Horse, Cattle, Extinction\\
200 & France, Italy, Spain, Latin, Netherlands\\
198 & Water, Agriculture, Coal, River, Antarctica\\
197 & Germany, World War II, Russia, World War I\\
187 & Mexico, Brazil, Atlantic Ocean, Argentina\\
185 & Human, Philosophy, Slavery, Religion, Democracy\\
184 & Plant, Rice, Fruit, Sugar, Wine, Maize, Cotton\\
177 & Gold, Iron, Oxygen, Copper, Electron, Color\\
170 & Egypt, Turkey, Israel, Islam, Iran, Middle East\\
159 & English, 19th century, William Shakespeare, Novel\\
158 & Africa, South Africa, Time zone, Portugal\\
157 & Europe, Scientific classification, Animal, Asia\\
141 & Washington, D.C., President of the United States\\
72 & Dinosaur, Fossil, Reptile, Cretaceous, Jurassic\\
70 & Paris, Art, Architecture, Painting, Hist. of painting\\
\hline
\end{tabular}
\caption{Global clustering of Wikipedia for Schools by weighted spectral embedding.}
\label{tab:global3}
\end{center}
\end{table}

The first observation is that the three embeddings are very different. The choice of the Laplacian (regular or normalized) matters, because the dimension $k$ is much smaller  than the number of nodes $n$ (recall that the shifted spectral embedding and the weighted spectral embedding are equivalent when $k = n$).  The second observation is that the weighted spectral embedding seems to better capture the structure of the graph. Apart from the first cluster, which is significantly larger than the others and thus may contain articles on very different topics, the other clusters look meaningful. The regular spectral embedding puts together articles as different as Meteorology, British English and Number, 
	while the shifted embedding groups together articles about Sanskrit, Light and The Simpsons, that should arguably appear in different clusters.

\clearpage

 \begin{table}[h]
\begin{center}
\begin{tabular}{|c|l|}
\hline
Size & Top articles\\
\hline
1666 & Germany, India, Africa, Spain, Russia, Asia\\
526 & Chordate, Binomial nomenclature, Bird, Mammal\\
508 & Earth, Water, Iron, Sun, Oxygen, Copper, Color\\
439 & Scotland, Ireland, Wales, Manchester, Royal Navy\\
300 & New York City, Los Angeles, California, Jamaica\\
246 & North America, German language, Rome, Kenya\\
195 & English, Japan, Italy, 19th century\\
131 & Language, Mass media, Library, Engineering, DVD\\
103 & Jazz, Piano, Guitar, Music of the United States\\
91 & Microsoft, Linux, Microsoft Windows, Algorithm\\
67 & British monarchy, Bristol, Oxford, Paul of Tarsus\\
58 & Tropical cyclone, Bermuda, Hurricane Andrew\\
49 & Mathematics, Symmetry, Geometry, Algebra, Euclid\\
48 & Fatty acid, List of vegetable oils, Biodiesel\\
47 & Train, Canadian Pacific Railway, Denver, Colorado\\
38 & Eye, Retina, Animation, Glasses, Lego, Retinol\\
27 & Finance, Supply and demand, Stock, Accountancy\\
19 & Meteorology, British English, Number, Vowel\\
17 & Newcastle upon Tyne, Isambard Kingdom Brunel\\
14 & Wikipedia, Jimmy Wales, Wikimedia Foundation\\
\hline
\hline
 Size & Top articles\\
\hline
452 & Chordate, Binomial nomenclature, Bird, Mammal\\
446 & UK, Europe, France, English language, Japan\\
369 & Germany, Spain, Soviet Union, Sweden, Poland\\
369 & Earth, Water, Iron, Sun, Oxygen, Copper, Color\\
353 & India, Africa, Russia, New Zealand, River, Snow\\
328 & Egypt, Greece, Middle Ages, Roman Empire, Nazism\\
286 & United States, New York City, Petroleum, Finland\\
271 & British Empire, 17th century, Winston Churchill\\
238 & Time zone, Turkey, Portugal, Israel, Currency\\
217 & Rice, Fruit, Bacteria, Wine, DNA, Flower, Banana\\
178 & British monarchy, Bristol, Charles II of England\\
175 & Internet, Hebrew language, Language, Mass media\\
156 & Physics, Ancient Egypt, Science, Astronomy, Time\\
125 & Atlantic Ocean, Morocco, Algeria, Barbados\\
124 & Opera, Folk music, Elvis Presley, Bob Dylan\\
117 & Agriculture, Ocean, Geology, Ecology, Pollution\\
112 & Christianity, Switzerland, Judaism, Bible, Deity\\
108 & South America, Pacific Ocean, Tourism, Colombia\\
94 & Film, Sanskrit, Light, The Simpsons, Eye, Shark\\
71 & Mining, Food, Geography, Engineering, Transport\\
\hline
\end{tabular}
\centering \caption{Global clustering of Wikipedia for Schools by regular (top) and  shifted (bottom) spectral embeddings.}
\label{tab:global1}
\end{center}
\end{table}

\vspace{.1cm}

\clearpage

 \paragraph*{Selective clustering.}
 We are now interested in the clustering of some selection of the articles. We take the 667 articles in the {\it People} category. A naive approach  consists in considering the subgraph induced by these nodes. But this is not satisfactory as the similarity between articles in the People category  depends strongly  on their connections through articles that not in this category. Indeed, the subgraph of nodes in the People category is not even connected. Our approach consists in assigning a multiplicative factor of 10 to articles in the People category. Specifically, we set 
   $w_i = 10 \times d_i$ if article $i$ belongs to the People category, and  $w_i = d_i$ otherwise. We compare the three previous embeddings, for this vector of weights.
   
     Tables \ref{tab:local3} and \ref{tab:local1} show the top  articles in the People category for the 20 clusters found for each embedding. The selected articles for each cluster correspond to the nodes of highest degrees in the People category. We also indicate the number of articles in the People category in each cluster.
Observe that the last  cluster obtained with  the  regular  spectral embedding  has no article in the People category. Again, the impact of weighting is significant. The weighted spectral embedding, which is adapted to the People category, seems to better capture  the structure of the graph. Except for the first, which is much larger than the others,  the clusters tend to group together articles of the People category that are closely related.

 \begin{table}[h]
\begin{center}
\begin{tabular}{|c|l|}
\hline
 Count & Top articles\\
\hline
228 & William Shakespeare, Pope John Paul II\\
60 & Jorge Luis Borges, Rabindranath Tagore\\
55 & Elizabeth II of the UK, Winston Churchill, Tony Blair\\
49 & Charles II of England, Elizabeth I of England\\
48 & Ronald Reagan, Bill Clinton, Franklin  Roosevelt\\
25 & Alexander the Great, Genghis Khan, Muhammad\\
24 & Adolf Hitler, Joseph Stalin, Vladimir Lenin\\
23 & Napoleon I, Charlemagne, Louis XIV of France\\
22 & Jesus, Homer, Julius Caesar, Virgil\\
22 & Aristotle, Plato, Charles Darwin, Karl Marx\\
21 & George W. Bush, Condoleezza Rice, Nelson Mandela\\
18 & Elvis Presley, Paul McCartney, Bob Dylan\\
15 & Isaac Newton, Galileo Galilei, Ptolemy\\
13 & Albert Einstein, Gottfried Leibniz, Bertrand Russell\\
10 & Pete Sampras, Boris Becker, Tim Henman\\
9 & William Thomson, 1st Baron Kelvin, Humphry Davy\\
8 & Carolus Linnaeus, James Cook, Gerald Durrell\\
8 & Bill Gates, Richard Stallman, Ralph Nader\\
5 & Floyd Mayweather Jr., Lucy, Lady Duff-Gordon\\
4 & Vasco da Gama, Idit Harel Caperton, Reza Shah\\
\hline
\end{tabular}
\caption{Selective clustering of Wikipedia for Schools by weighted spectral clustering.}
\label{tab:local3}
\end{center}
\end{table}

 \begin{table}[h]
\begin{center}
\begin{tabular}{|c|l|}
\hline
Count & Top articles\\
\hline
196 & Carolus Linnaeus, Adolf Hitler, Jesus, Aristotle\\
95 & Christina Aguilera, Andy Warhol, Auguste Rodin\\
90 & Julius Caesar, Martin Luther King, Jr., Euclid\\
75 & Tony Blair, Victoria of the UK, Charles II of England\\
45 & Ronald Reagan, Franklin D. Roosevelt, Gerald Ford\\
28 & Abraham Lincoln, George Washington, John Adams\\
25 & Igor Stravinsky, Johann Sebastian Bach\\
24 & Albert Einstein, Gottfried Leibniz, Isaac Newton\\
22 & George W. Bush, Napoleon I of France, Plato\\
10 & Pete Sampras, Boris Becker, Tim Henman, Pat Cash\\
10 & Fanny Blankers-Koen, Rosa Parks, Donald Bradman\\
9 & Alexander the Great, Frederick II of Prussia\\
9 & Mahatma Gandhi, Buddha, Muhammad Ali Jinnah\\
8 & Columba, Edwin of Northumbria, Macbeth of Scotland\\
5 & Bill Clinton, Richard Stallman, Linus Torvalds\\
5 & Elizabeth II of the UK, David Beckham, Wayne Rooney\\
5 & Archbishop of Canterbury, Harold Wilson\\
4 & Leonardo da Vinci, Neil Armstrong, Wright brothers\\
2 & George III of the UK, Matthew Brettingham\\
0 & $-$\\
\hline
 Count & Top articles\\
\hline
182 & Adolf Hitler, William Shakespeare\\
113 & Elizabeth II of the UK, George W. Bush\\
69 & Tony Blair, Victoria of the UK, Elizabeth I of England\\
46 & Ronald Reagan, Franklin Roosevelt, Jimmy Carter\\
31 & Henry James, Igor Stravinsky, Ezra Pound\\
26 & Paul McCartney, Bob Dylan, Edgar Allan Poe\\
25 & Jesus, Charlemagne, Genghis Khan, Homer, Ptolemy\\
24 & Albert Einstein, Gottfried Leibniz, Isaac Newton\\
23 & Charles Darwin, Galileo Galilei, Nikola Tesla\\
19 & Plato, John Locke, Max Weber, Friedrich Nietzsche\\
18 & Rabindranath Tagore, Mahatma Gandhi, Buddha\\
17 & Margaret Thatcher, David Cameron, George VI\\
16 & Condoleezza Rice, Nelson Mandela, Gerald Ford\\
12 & Dwight D. Eisenhower, Ernest Hemingway\\
11 & Aristotle, Alexander the Great, Fred. II of Prussia\\
10 & Pete Sampras, Boris Becker, Tim Henman\\
8 & Muhammad, Norman Borlaug, Osama bin Laden\\
7 & Bill Clinton, Bill Gates, Richard Stallman\\
5 & Leonardo da Vinci, Neil Armstrong\\
5 & Carolus Linnaeus, Christopher Columbus, Paul Kane\\
\hline
 \end{tabular}
\caption{Selective clustering of Wikipedia for Schools by regular (top) and shifted (bottom) spectral clustering.}
\label{tab:local1}
\end{center}
\end{table}

\clearpage

%

\section{Conclusion}
 
 \label{sec:conc}

We have proposed a novel spectral embedding of graphs that takes node weights into account. We have proved that the dimensions of this embedding correspond to different configurations of an equivalent physical system, either mechanical or electrical, with node weights corresponding to masses or capacitances, respectively.

A practically interesting consequence of our work is in the choice of the Laplacian, when there are no other information on the node weights than the graph itself. Thanks to weighted spectral embedding, we see that the two versions of the Laplacian, regular and normalized, correspond in fact to two relative importance of nodes, given respectively  by unit weights and internal node weights. 

\end{document}